# CLASSIFICATION OF QUESTIONS AND LEARNING OUTCOME STATEMENTS (LOS) INTO BLOOM'S TAXONOMY (BT) BY SIMILARITY MEASUREMENTS.

## TOWARDS EXTRACTING OF LEARNING OUTCOME FROM LEARNING MATERIAL

Shadi Diab[1] and Badie Sartawi[2]

[1]Information and Communication Technology Center, Al-Quds Open University, Ramallah – Palestine

[2]Associate Professor of Computer Science, Al-Quds University, Jerusalem - Palestine

*ABSTRACT*

*Bloom's Taxonomy (BT) has been used to classify the objectives of learning outcome by dividing the learning into three different domains; the cognitive domain, the affective domain, and the psychomotor domain. In this paper, we introduced a new approach to classify the questions and learning outcome statements (LOS) into Bloom's taxonomy (BT) and to verify BT verb lists, which are being cited and used by academicians to write questions and (LOS). An experiment was designed to investigate the semantic relationship between the action verbs used in both questions and LOS to obtain a more accurate classification of the levels of BT. A sample of 775 different action verbs collected from different universities allows us to measure an accurate and clear-cut cognitive level for the action verb. It is worth mentioning that natural language processing techniques were used to develop our rules to induce the questions into chunks in order to extract the action verbs. Our proposed solution was able to classify the action verb into a precise level of the cognitive domain. We, on our side, have tested and evaluated our proposed solution using a confusion matrix. The results of evaluation tests yielded 97% for the macro average of precision and 90% for F1. Thus, the outcome of the research suggests that it is crucial to analyze and verify the action verbs cited and used by academicians when writing LOS and when classifying their questions based on blooms taxonomy in order to obtain a definite and more accurate classification.*

## 1. INTRODUCTION

The new international trends in education show a shift from traditional teacher-centered approach to a "student-centered" approach, which focuses, in turn, on what the students are expected to do at the end of the learning process! Therefore, this approach is commonly referred to as an outcome-based approach. Statements called intended learning outcomes, commonly shortened to learning outcomes, are being used to express what the students are expected to be able to do at the end of the learning period [1]. Learning is defined, in term of its outcome, in different contexts and for different purposes or settings e.g. in terms of education, work, guidance and personnel context [2]. As for our research, it focuses on the education context presented in the form of textbooks deployed by the teaching staff. Learning outcomes can be defined for a single course taught by several teachers, or be standardized across universities or whole domains. Instructional designers (represented by the author of the textbook itself) should be provided a list of relevant learning outcome definitions they can link to their courses [3]. There are many useful guides for developing a comprehensive list of student outcomes. For example, Bloom's taxonomy is used to define the objective of learning and teaching as well as to divide learning into three types of domains: cognitive, affective and psychomotor. Then, it defines the level of performance for each domain [4]. Former students of blooms and a group of cognitive psychologists, curriculum theorists, and instructional researchers have released a new version of Bloom's taxonomy in 2001 [5]. Our research will focus on the cognitive domain of Bloom's Taxonomy 1.

It is a truism for educators that questions play an important role in teaching [6]. Our research focuses on questions classification into a cognitive level of Bloom's taxonomy, which is a framework for classifying educational goals and objectives into a hierarchical structure representing levels of learning. BT is of three different domains: the cognitive domain, the affective domain, and the psychomotor domain. Each of these has a multi-tiered hierarchical structure for classifying learning [5]. The Cognitive Domain (Bloom et al., 1956) has become widely used throughout the world to assist in the preparation of evaluation materials [1].

There are six major categories (levels). The levels are Knowledge; Comprehension; Application; Analysis; Synthesis and Evaluation [7]. In our proposed approach, we will use the action verb of the question or (LOS) which represents the cognitive skill to classify the question into one or more levels.

## 2. LITERATURE AND RELATED WORK

Many researchers attempted to classify questions into different classes and for different purposes. In [8] they classified learning questions through a machine learning approach and learned a hierarchical classifier guided by a layered semantic hierarchy of answer types. They eventually classified questions into fine-grained classes. Their hierarchal classifier achieved 98.80% precision for coarse classes with all the features, and 95% for the fine classes.

Keywords database matching with the verb of the question method has been developed, piloted and tested for automatic Bloom's taxonomy analysis, that matches all levels of cognitive domain of bloom [9], the results have shown that the knowledge level achieved 75% correct match in comparison with the expert's results. their system allows both teachers and students to work together in the same platform to insert questions and review learning-outcome matches with the cognitive domain of BT.

[10] They proposed natural language processing-based automatic concept extraction and outlines rule-based approach for separation of prerequisite concepts and learning outcomes covered in learning document, by their manual creation of domain ontology. Their system achieved Precision: 0.67 Recall: 0.83 F-score: 0.75.

[11] They also proposed a rule-based approach to analyze and classify written examination questions through natural language processing for computer programming subjects, the rules were developed using the syntactic structure of each question to apply the pattern of each question to the cognitive level of bloom. Their evaluation achieved macro F1 of 0.77. The researchers, in their other research, [12] proposed *Bloom's Taxonomy Question Categorization Using Rules and N-Gram Approaches*. In their experiment; 100 questions were selected for training and 35 were used for testing and both were based on programming domain. The categorization uses a rule-based approach, N-gram and a combination of both. Their result demonstrated that combination rule-based and n-gram approaches obtained the highest and the best score of precision of an average of 88%.

[13] researchers have taken data of Li and Roth in [8] to classify the questions into three broad categories instead of 6-course grain and 50 fine-grained categories. They analyzed the questions syntactically to expect the answer type for every particular category of the questions. [14] They

also classified questions with different five machine learning algorithms: Nearest Neighbours (NN); Naïve Bayes (NB); Decision Tree (DT); Sparse Network of Windows (SNoW); and Support Vector Machines (SVM). They did the classification using two features: bag-of-words and bag-of n-grams. They proposed a special kernel function to enable (SVM) take advantages of the syntactic structure of the questions. In their experiment, the questions classification accuracy reached 90%.

[15] They proposed two Level Question Classification based on SVM and Question Semantic Similarity in computer service & support domain, their results showed that question classification dramatically improves when complementing the domain ontology knowledge with question-specific domain concepts. They also presented a two-level classification approach based on SVM and question semantic similarity. [16] They also explored the effectiveness of support vector machines (SVMs) to classify questions, their evaluation showed the micro was 87.4 accuracy, 83.33 precision, and 44.64 F1.

Most of the researchers in our literature review had focused on classifying questions into different classes, including the classes of cognitive levels of BT... Purely machine learning and rules-based approaches have been applied. Most of these researchers used a huge amount of data and domain ontology to run their experiments, including the need to domain-experts to evaluate the performance, we consider [17] is the most related research to our approach. They used WordNet with cosine algorithm to classify exams question into bloom taxonomy. Questions pattern identification was required as a step to measure the cosine similarity by finding the total number of WordNet values for questions and run cosine similarity twice; first for pattern detection and second after calculating the WordNet value. Their evaluation achieved 32 questions out of 45 correctly classified. However, in our research, we proposed one similarity algorithm to measure the semantic similarity between the action verb of the question and the action verb list categorized by domain experts to find out the most accurate level for the question. Moreover, our algorithm was evaluated using a confusion matrix. It was applied to both the cognitive domain of BT and the remaining two domains.

## 3. SEMANTIC SIMILARITY MEASUREMENT

### 3.1 . Semantic Similarity

The semantic similarity has attracted great concern for a long time in artificial intelligence, psychology and cognitive science. In recent years, the measures based on WordNet have shown its capabilities and attracted great concern [18]. Researchers used a measure of semantic relatedness to perform the task of word sense disambiguation [19]. Semantic similarity measures can be generally partitioned based on four grounds: based on the distance similarity between two concepts; based on information the two concepts share; based on the properties of the concepts; and based on a combination of the previous options [20].

### 3.2 Wordnet

WordNet is a large lexical database of English. Nouns, verbs, adjectives, and adverbs are grouped into sets of synonyms (Synsets). Synsets are interlinked by means of conceptual-semantic and lexical relations. It includes 82115 nouns, 13767 verbs, 18156 adjectives, 3621 adverbs [21]. Wu and Palmer (Wu and Palmer, 1994) similarity metric measure semantic similarity through the depth of the two concepts in the WordNet taxonomy [22]. However, there are some important distinctions: First, WordNet interlinks not just word forms strings of letters but also specific senses of words. As a result, words found in close proximity to one another in the network are semantically disambiguated. Second, WordNet labels the semantic relations among words, whereas the groupings of words in a thesaurus do not follow any explicit pattern other than meaning similarity [22]. Wu-Palmer representation scheme does not only take care of the semantic-syntactic correspondence, but it also provides similarity measures for the system for the performance of inexact matches based on verb meanings [23]. The wu-palmer algorithm uses the following equation to measure the similarity:

$$\frac{2 * \text{depth(LCS)}}{\text{depth(concept1)} + \text{depth(concept2)}}$$

## 4. RESEARCH'S METHODOLOGY

Analyzing questions and LOS to determine the most accurate level in BT domains is a challenge. This will lead us to discover the intended learning outcome that will be achieved by the students. In our research, we concentrated on the action verbs that should be used to write questions and LOS based on the cognitive domain by analyzing the questions and LOS.

We have observed that categorization of the actions verbs may occur in different levels of the cognitive domain, thus, you may find the verb *write* in knowledge, application, comprehension or analysis levels, such this classification depends on the understanding of the action verb classified by domain experts. Academicians would manually classify the question into a taxonomy level based on their styles [11]. Through our research, we will answer the following questions:

How can we classify the question and LOS into one or more a level of the cognitive domain using semantic similarity measurements? Does our proposed approach apply to the two remaining domains of BT? Will semantic similarity between action verbs of the question and the action verb lists assist in the enhancement of classifying questions and the writing of more accurate LOS?

## 5. COLLECTING DATA FROM DOMAIN EXPERTS

We have observed that many universities, worldwide, prepared guidelines and specific publications to support their teachers in writing questions and LOS. such instructions guides point to specifics action verbs as a reference to classify the verbs into BT. By assuming that the teachers use guides and supportive publications of their schools and universities in order to write questions and LOS, We collected 605 different action verbs that describe the cognitive skills in each level from websites of different universities [24] [25] [26] [27]. To gain more accurate and precise data, we filtered and modified the data lists by collecting the verbs intersecting with three or four lists (threshold 75-100%). We also added verbs intersecting with two resources if and only if having no conflict with other lists (threshold 50%). The result was a new dataset that contains 77 different action verbs distributed on the six levels of the cognitive domain of BT. Moreover, *questions starters* from [28], which organize the starters of questions that cover each level of the cognitive domain of BT, has been collected.

## 6. STRUCTURAL INDUCTION OF THE QUESTION

Structural induction may be defined as the process of extracting structural information using machine learning techniques and the patterns found may use to classify the questions [29]. This allows us to take some parts of the question and leave the others for further processing. Our experiment aims to extract the action verb of the question by using structural induction. Using the questions starters collected from [28], we were able to extract the action verb of the questions throughout implementing the following steps:
splitting the questions into separate lines, tokenization, lemmatization, POS tagging, partial parser over grammar which detect main action verb of the question, we were able to convert such question in form of POS tags patterns contain the action verb of the question.

For example, running partial parsing over manually built in grammar to detect the knowledge level of the cognitive domain based on starters of [28]: Q: How would you explain computer science to a five-year-old? Steps will return the chunked tree labeled with "KNOW" as in Figure 1, while the main action verb *explain* refers to the knowledge level of BT

*Figure 1: Chunked Tree Example*

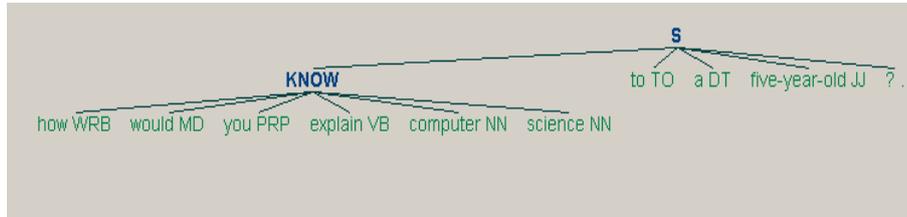

We have observed, after the implementation of our experiment, that adapting partial parser over built-in grammars is applicable and effective to extract the action verb in order to move forward in our next experiments and analysis.

## 7. THE PROPOSED ACTION VERBS CLASSIFICATION ALGORITHM

Different verbs can be used to demonstrate different levels of learning, for example, the basic level the learning outcomes may require learners to be able to define, recall, list, describe, explain or discuss [2]. In addition to that, the verb is considered the center, the fulcrum and the engine of a learning outcome statement. We should note that verbs refer to events, not to states; events are specific actions [30]. Thus, our proposed solution is based on the classification of action verb of the questions or LOS, in order to classify the whole question or LOS into a more accurate level. The following definitions and steps describe our algorithm:

BTD (Bloom's Taxonomy Dimensions) = [C, A, P] where denotes for cognitive, affective and the psychomotor domains respectively.
Based on BT classification each dimension of BT contains different levels (L), where the cognitive domain (C) contains six levels, and each affective (A) and psychomotor (P) domain contains 5 levels, thus C= [L1...L6], A= [L1... L5], and P= [L1... L5], for each level (L) in any dimension there are some groups of action verbs represent the particular level, these verbs assist and support the academicians to write LOS and questions based on BT.

Classification of the action verb of the question or LOS (VQ) into one or more of dimension of BTD, and by similarity measurement between the action verb in VQ and each verbs of L in C, A or P by calculating similarity (sim) measurements, maximum similarity ($Max_{sim}$), and the total of similarities in each level, our algorithm will find three main measurements as follow:

- Similarity measurement between the obtained action verb of the question or LOS and each of verbs represent the level of BT, Sim = Similarity_algorithm (VQ, N), where N is number of verbs represent the particular level.
- Maximum similarity value between the action verb of the question or LOS and the verbs represent each level of BT, $Max_{sim}$ = Maximum of semantic similarities between (VQ, N) for each L.

- The maximum area represents the total amount of similarity values for the action verb of the question or LOS in each level, $Max_{area}$ of L in C, A or P = MAX ( $\sum_{i=0}^{n} Sim(i)$ = sim (0) + sim (1) +… + sim (n), where i >=0 and n is the list of action verbs.

## 8. ACTION VERBS CLASSIFICATION ALGORITHM (AVCA)

For the sake of simplicity, our algorithm and implementation applied on the cognitive domain, while the data (verbs) represent the cognitive domain are same type of data represent the other domains but with different verbs, furthermore our algorithm may accept any input data in form of verbs regardless if its related to cognitive, affective or psychomotor verbs, the proposed algorithm Pseudo code and step as follow:

**(Pseudocode):**
**Algorithm AVCA (VQ [0...M], CL [1...N], $Max_{sim}$ [1…6], $Max_{area}$):**

// the algorithm measure the similarity between groups of verbs
//by calculating the high similarity and total amount of similarity values
// Input: List of action verbs obtained from questions or LOS, VQ [1...M], and list of verbs
//represent the cognitive domain CL, where CL= [L1...L6] and each contains a group of verbs
L= //[1...N]
//Output: The maximum similarity between each verb of VQ list and L in CL, and maximum similarity area for each verb of VQ in each L in CL

For each L in CL:

    Compute Sim = [Similarity_algorithm (M, N)]
    Compute $Max_{sim}$ = [Max (sim)]
    Classification result = L with greatest (Max (sim))

    IF Len ($Max_{sim}$) >1 //have more than one max similarity appears in more than one level L
      For each L in CL:
        $Max_{area}$ = Sum (sim1, sim2….Sim...n)
        Classification result = L with greatest ($Max_{area}$)
    STOP

While there are a few similarity algorithms adapted in WORDNET, we implemented our algorithm on Wu-Palmer similarity algorithm to measure the similarity values, maximum similarities and similarity area, in additional, our algorithm will remain correct and applicable on the other similarity algorithms, and while the input data types are all in form of action verbs regardless if belong to cognitive, affective or psychomotor, our algorithm also will generate correct results and remain applicable for any dimension of BT.

## 9. EXPERIMENT AND ANALYSIS

Our classification algorithm applied the constructed verb lists from questions and LOS to compute the maximum similarity for each level of the cognitive domain. Then it compares the maximum similarities to nominate (the greater) one and only one level as an accurate level for the classified verb. Our experiment was built based on the collected data from [24] [25] [26] [27]. Such data has been built to support the academicians to write questions and LOS, we observed the following behaviors and cases:

Identical similarity will appear when the Synsets of the action verb has 1 similarity value with one or more verbs in the lists of cognitive domain. For example, figure 2 shows that the verb *compile* has 1 similarity value with the verb roll up in the synthesis level. We may conclude that the verb compile is way closer to the synthesis level than the other levels.

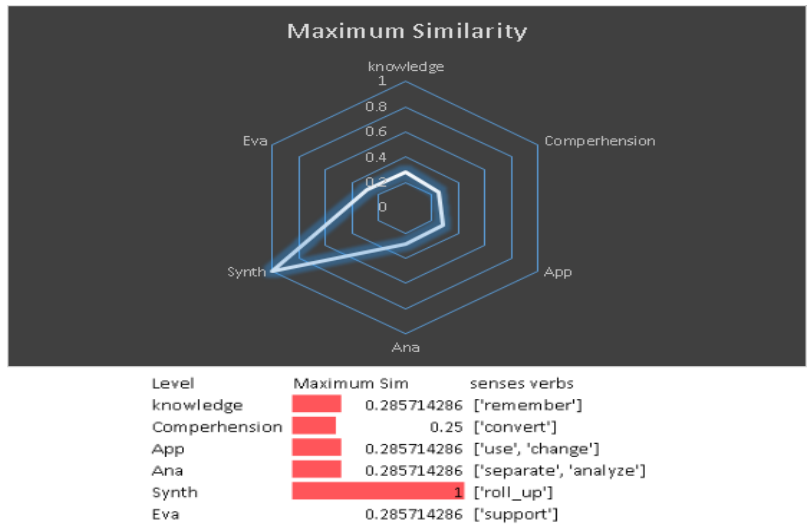

*Figure 2: Maximum Similarity for the verb <compile>*

Higher similarity value will appear when the action verb has less than 1 maximum similarity value. It also appears in one and only one level of the cognitive domain. For example, figure 3 shows that the verb *write* has a higher similarity value (0.857) with the verb *dramatize* in the application level. Thus, we may conclude that the verb dramatize belongs to application level more than the others do.

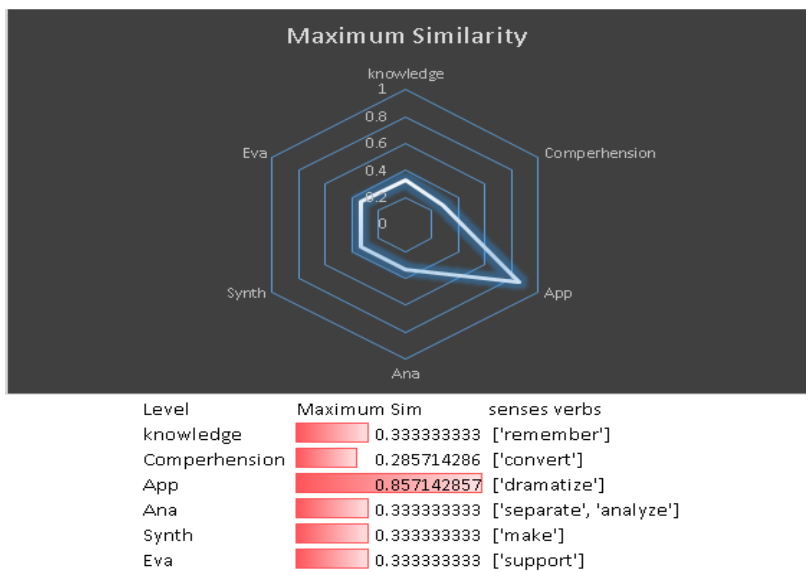

*Figure 3: Maximum similarity for the verb <write>*

Obtaining same maximum similarity for some action verbs in more than one cognitive level may mean that the verb of the question could be applied to more than one level, see figure 4,

Such case, in the point of view of some academicians may make sense. Moreover, we could prove that the action verb of the question may belong to one and only one level of the cognitive domain and have greater similarity semantically than the others. Figure 4 shows that the verb manipulates have a maximum similarity (0.28) in all levels.

*Figure 4: Maximum Similarity for the verb < manipulate >*

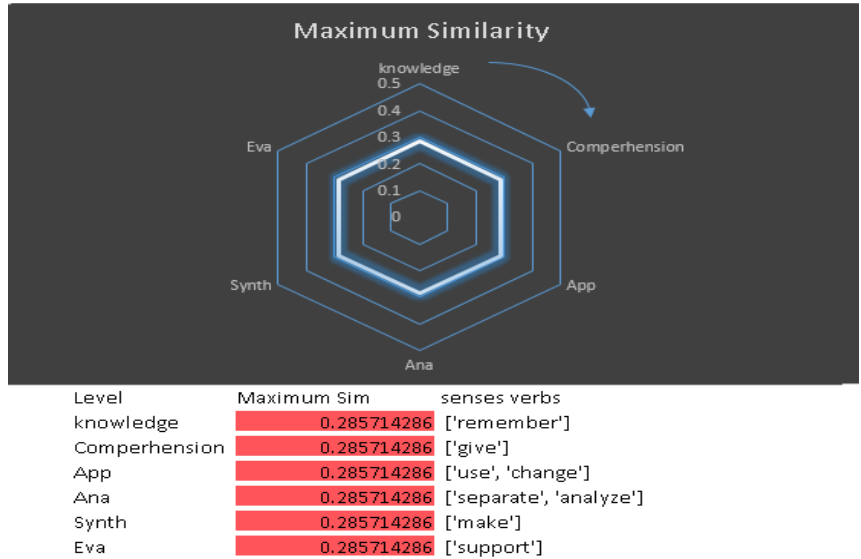

In order to check if the similarity values have a bias to one level more than the others. We calculated the similarity values for each one-sixth. Figure 5 shows that the bias will appear in application level where the total amount of all similarities values in application level is greater than the others. In all experiments we conducted, we were not able to find two verbs of the same total similarity values in more than one-sixth. This leads us to measure and classify the verb in one and only one level of the cognitive domain of BT.

*Figure 5: Computing of similarities in each one six*

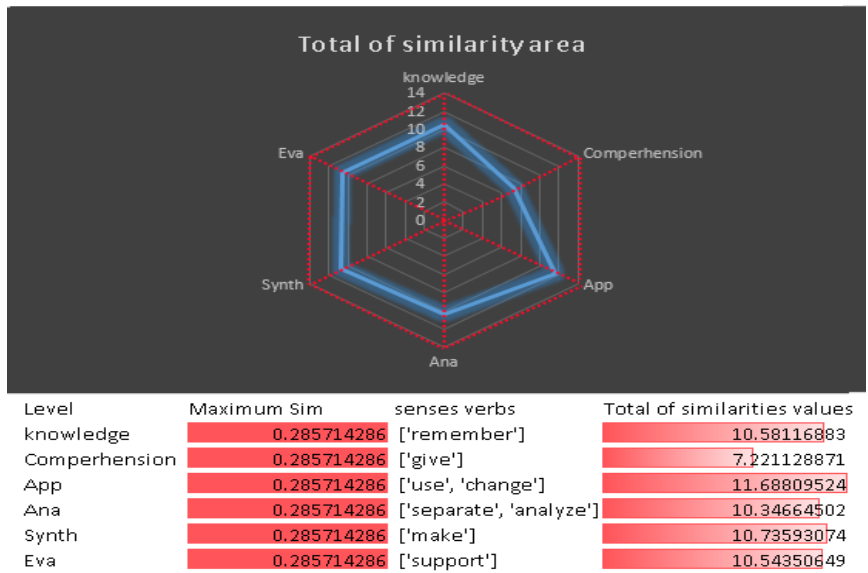

## 10. ENHANCEMENT OF THE ACTION VERBS LIST

We validate our proposed algorithm on new data sets of action verbs collected from different resources from [31] [32]. All verbs were tested against each verb in our collected data [24] [25] [26] [27]. We found that our proposed algorithm can improve the correctness of the categorization based on cognitive blooms taxonomy from an average of 71% to 97%.

However, our algorithm was able to find that (34%) were incorrectly manually classified, on its part, has reduced the incorrectly classified verbs percentage from 34% to 8%. The improvement average detected was 97% for both data sets as a result of applying a threshold of ≥ 50% as evidence from our source data.

## 11. EVALUATION AND RESULTS

We evaluated our algorithm to measure the performance and our obtained results. We used a confusion matrix, which is often used to describe the performance of the classification model [33] in order to measure the following values:

- True Positives (TP) is the count of correctly predicted positive values. We count the number of TP for each verb in our implementation as if the actual verb is classified in the same class or level (both of the two verbs belong to the same cognitive level)

- True Negatives (TN) is the count of correctly predicted negative values; the total amount of cases when the result of classification is false and the actual verb is also in the different class. (Both of our predicted result and the actual classified verb are not in the correct level.)

- False Positives (FP): when the prediction result is yes and the actual verb is in a different class.

- False Negatives (FN): when the prediction return negative result but the actual is true.

Evaluation of our algorithm was based on unseen data collected from [31] [32]. Moreover, a threshold of ≥ 50% used to measure the actual level of each verb in [24] [25] [26] [27]. By comparing the result and the actual level for each verb, we were able to create the confusion matrix in order to measure the most common measures used to evaluate the performance:

- Accuracy: The simplest metric that can be used for evaluation; it measures the percentage of inputs in the test set that the classifier correctly labeled [33]. it also can be measured by calculating TP+TN/TP+FP+FN+TN [34]
- Precision: Indicates how many of the items that we identified were relevant and can be measured by calculating TP/ (TP+FP) [33].
- Recall: Indicates how many of the relevant items that we identified, and measured by TP/ (TP+FN) [33].
- F1 (or F-Score): combines the precision and recall to give a single score. F1 is defined to be the harmonic mean of the precision and recall and measured as follow:
    (2 × Precision × Recall)/ (Precision Recall) [33].
- Error-Rate (ERR): the calculated number of all incorrect prediction divided by the total number of the dataset = FP + FN / N [35]

- Macro-Average: To calculate the harmonic mean of precision, recall for all classes (levels). The obtained results after processing the test sets in [31] and [32] summarized in table 1 and table 2.

*Table 1: Evaluation Summary for processing dataset 1*

| Level / Measures | Accuracy | Precision | Recall | F1 | Error Rate |
|---|---|---|---|---|---|
| **Knowledge** | 91% | 100% | 0.90909909 | 0.952380952 | 0.09090909 |
| **Comprehension** | 69% | 100% | 0.69237692 | 0.818181818 | 0.30769231 |
| **Application** | 71% | 91% | 0.76923769 | 0.833333333 | 0.28571429 |
| **Analysis** | 93% | 92% | 1 | 0.956521739 | 0.07142857 |
| **Synthesis** | 76% | 100% | 0.73684215 | 0.848484848 | 0.23809524 |
| **Evaluation** | 100% | 100% | 1 | 1 | 0 |
| **Macro-Average** | **83%** | **97%** | **85%** | **90%** | **0.16563992** |

*Table 2: Evaluation Summary for processing dataset 2*

| Level / Measures | Accuracy | Precision | Recall | F1 | Error rate |
|---|---|---|---|---|---|
| **Knowledge** | 91% | 100% | 91% | 0.952380952 | 0.09090909 |
| **Comprehension** | 69% | 100% | 67% | 0.8 | 0.30769231 |
| **Application** | 93% | 100% | 93% | 0.962962963 | 0.07142857 |
| **Analysis** | 93% | 75% | 100% | 0.857142857 | 0.07142857 |
| **Synthesis** | 76% | 93% | 76% | 0.838709677 | 0.23809524 |
| **Evaluation** | 75% | 100% | 73% | 0.842105263 | 0.25 |
| **Macro-Average** | **83%** | **95%** | **83%** | **0.875550286** | **0.1715923** |

It can be concluded that the accuracy, precision, and recall in each level in our two evaluation tests are very satisfactory. Moreover, our algorithms' evaluation overall is very satisfactory as well. In both our validation and evaluation, we were able to prove that the classification of the action verb into one or more level of cognitive domain of BT will increase the efficiency of such classification. This can be used to enhance not only writing learning outcome statements but also classifying the question into a more accurate level semantically.

## 12. CONCLUSION

We introduced the classification problem of the questions and LOS into bloom taxonomy. Our research explored the rules-based approach to induct the most important part of the question. Such parts, which include the action verb of the question, will lead us to measure the accurate level of the action verb in the cognitive domain. We also conducted an analysis of currently used action-verb lists as guides and manual instructions for academicians and proposed a new method to measure the relationship between these verbs, the verb of the question and the learning outcome statements (LOS). We, as well, adapted similarity measures to provide an accurate classification for such verbs by two methods; using the maximum similarity and calculating the whole similarity area for each one six of the cognitive domain hierarchy. We have validated and proved that our proposed solution will improve the classified action verbs into more accurate levels. Later, we evaluated our proposed method by using the confusion matrix and measured a very high Macro-average of precisions for all one-six of the cognitive domain of BT. In conclusion, this will enhance the cognitive action verb lists. These lists, however, are being used and cited by academicians to write LOS and classify their questions based on blooms taxonomy as this work helps provide more accurate verbs, this will, in turn, provide more accurate intended mental skills. This will also, in addition to the previously mentioned, clear the ambiguity that lies behind the classification of questions into bloom taxonomy as well as the action verbs when used in writing LOS.

## 13. FUTURE WORK

Deeply Syntax analysis is required to convert the whole question into LOS automatically. Moreover, analyzing the figures such as images, graphs, and tables in order to construct LOS, which represents the objectives behind by such figures, is very important toward constructing LOS from learning material.

## REFERENCES

[1]     Declan Kennedy, Writing and using learning outcomes: a practical guide, Page 18, Cork: University College Cork, 2006.

[2]     EU, Publications Office, Using learning outcome, European Qualifications Framework Series: Note 4, Luxembourg: ISBN 978-92-79-21085-3, 2011 Page, 2010.

[3]     Totschnig, Michael, et al., "Repository services for outcome-based learning" 2010.

[4]     Office Of Institutional Effectiveness, Assessment Workbook For Academic Programs, Richmond: University Of Richmond, 2008.

[5]     Advancing Academic Quality in Business Education Worldwide, Bloom's Taxonomy of Educational Objectives and Writing Intended Learning Outcomes Statements, Kansas: international Assembly for Collegiate Business Education, 2016.

[6]     M. D. Gall, "The Use of Questions in Teaching," American Educational Research Association, vol. 40, no. 5, pp. 707-721, 1970.

[7]     D. R. Krathwohl, A Revision of Bloom's Taxonomy:, College of Education, The Ohio State University.

[8]     Li and Roth, Learning Question Classifiers, Illinois USA: 19th International Conference on compuatational Linguistics (COLING) 2002.

[9]     Wen-Chih Chang and Ming-Shun Chung, Automatic Applying Bloom's Taxonomy to Classify and Analysis the cognition level of English questions items, Taiwan: IEEE, 2009.

[10]    Sonal Jain and Jyoti Pareek, Automatic extraction of prerequisites and learning outcome from learning material, India: Inderscience Enterprises Ltd., 2013.

[11]    Syahidah Haris and Nazila Omar, A Rule-based Approach in Bloom's Taxonomy Question Classification through Natural Language, Malaysia, 2013.

[12]    Syahidah sufi haris and nazlia omar, Bloom's Taxonomy Question Categorization Using Rules And N-Gram Approach, Malaysia: Journal of Theoretical and Applied Information Technology, 2015.

[13]    Payal Biswas et al, Question Classification Using Syntactic And Rule-Based Approach, New Delhi: International Conference on Advances in Computing, Communications, and Informatics (ICACCI), 2014.

[14]    Dell Zhang et al, Question Classification using Support Vector Machines, Singapore: ACM 1-58113-646-3/03/0007, 2003.

[15]    Jibin Fu, Two Level Question Classification Based on SVM and Question Semantic Similarity, Beijing, China: International Conference on Electronic Computer Technology, 2009.

[16]    Anwar Ali Yahya and Addin Osman, Automatic Classification Of Questions Into Bloom's Cognitive Levels Using Support Vector Machines, Najran: Researchgate, 2011.

[17]    K. J. et al, An automatic classifier for exam questions with wordnet and cosine, International Journal of Emerging Technologies in Learning (iJET) 11(04):142 · April 2016.

[18]     Lingling Meng et al, "A Review of Semantic Similarity Measures in WordNet," International Journal of Hybrid Information Technology, vol. 6, no. 1, p. 1, 2013.

[19]     Siddharth Patwardhan et al, "Using Measures of Semantic Relatedness for Word Sense Disambiguation," Springer, p. 241–257, 2003.

[20]     G. V. e. al, "Semantic Similarity Methods in WordNet and Their Application to Information Retrieval on the Web," ACM New York, NY, USA ©2005, pp. 10-16, 2005.

[21]     G. A. Miller, "WordNet - A Lexical Database for English," Communications of the ACM, vol. 38, no. 11, pp. 39-41, 2005.

[22]     C. C. a. R. Mihalcea, Measuring the Semantic Similarity of Texts, University of North Texas: University of North Texas, 2005.

[23]     Z. W. a. M. Palmer, "VERB SEMANTICS AND LEXICAL SELECTION," in In Proceedings of the 32nd Annual Meeting of the Associations for Computational Linguistics, Las Cruces, New Mexico,, 1994.

[24]     Virginia.ED, "Bloom's Taxonomy of Cognitive Skills with Action Verb List Source," University of Virginia, 20 2 2017. [Online]. Available: http://avillage.web.virginia.edu/iaas/assess/resources/verb-list.pdf. [Accessed 20 2 2017].

[25]     U. O. C. Florida, "Bloom's Taxonomy," University of Central Florida, 20 2 2017. [Online]. Available: http://www.fctl.ucf.edu/teachingandlearningresources/coursedesign/bloomstaxonomy/. [Accessed 20 2 2017].

[26]     M. S. University, "Bloom's Taxonomy Action Verbs," Missouri State University, 20 2 2017. [Online]. Available: https://www.missouristate.edu/assets/fctl/Blooms_Taxonomy_Action_Verbs.pdf. [Accessed 20 2 2017].

[27]     C. A. College, "Bloom's Taxonomy Verbs," Central Arizona College, 20 2 2017. [Online]. Available: http://www.centralaz.edu/Documents/class/Bloom's%20Taxonomy%20Verbs%20Web.pdf. [Accessed 20 2 2017].

[28]     F. Kugelman, "http://www.bloomstaxonomy.org/," 6 2 2017. [Online]. Available: http://www.bloomstaxonomy.org/BloomsTaxonomyquestions.pdf. [Accessed 06 02 2017].

[29]     Zaanen et al, "Question classification by structure induction," in IJCAI'05 Proceedings of the 19th international joint conference on Artificial intelligence, Edinburgh, Scotland, 2005.

[30]     C. Adelman, The Language, and Syntax of Learning Outcomes Statements, National Institute for Learning Outcomes Assessment, 2015.

[31]     The Center For Learning, Muskie Institute Center for Learning, University of Southern Maine [Online]. Available: http://php.ipsiconnect.org/doc/VerbList.htm. [Accessed 27 2 2017].

[32]     Fresnostate university, "Bloom's Taxonomy Action Verbs," [Online]. Available: http://www.fresnostate.edu/academics/oie/documents/assesments/Blooms%20Level.pdf. [Accessed 4 3 2017].

[33]     K. Markham, "Simple guide to confusion matrix terminology," dataschool, [Online]. Available: http://www.dataschool.io/simple-guide-to-confusion-matrix-terminology/. [Accessed 07 03 2017].

[34]     E. K. A. E. L. Steven Bird, Natural Language Processing with Python, 2009: O'Reilly Media, Inc., O'Reilly Media, Inc.

[35]     R. Joshi, "How to evaluate the performance of a model in Azure ML and understanding "Confusion Metrics"," exsilio.com, 9 September 2016. [Online]. Available: http://blog.exsilio.com/all/accuracy-precision-recall-f1-score-interpretation-of-performance-measures/. [Accessed 2017 3 7].

## AUTHORS

**Shadi Diab** received his M.Sc. degree in computer science from Al-Quds University - Jerusalem - Palestine in 2017, and his B.Sc. in computer information systems from Al-Quds Open University, since 2008, he is head of accreditations and internet-based testing unit in Information and Communication Technology Center (ICTC) in Al-Quds Open University - Palestine

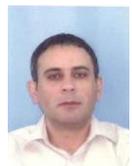

**Badie Sartawi** received his Ph.D. in Systems Theory and Engineering with major in Computer Engineering, University of Toledo, Ohio. 1993, Associate Professor at CS & IT department of Al-Quds University - Jerusalem - Palestine for the past 20 years.  He has been involved in the design of various education and professional ICT programs and he is among that active and visionary person in the education and ICT community.

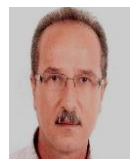